%% file: main.tex
\pdfoutput=1

\documentclass[11pt]{article}

\usepackage[]{ACL2023}

\usepackage{times}
\usepackage{latexsym}

\usepackage[T1]{fontenc}

\usepackage[utf8]{inputenc}

\usepackage{microtype}

\usepackage{inconsolata}

\usepackage{graphicx}
\usepackage{float}
\title{\mymodel{}: Contrasting Large and Small Language Models to \\ Verify Grounded Generation \\  \includegraphics[width=3cm]{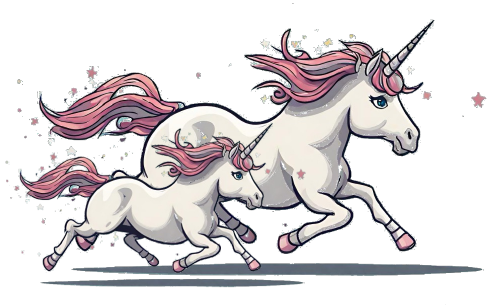}}

\author{
I-Hung Hsu$^{1}$\thanks{\ Work done while the author was a student researcher at Google Cloud AI Research. Correspondence to: I-Hung Hsu <ihunghsu@usc.edu>, Chen-Yu Lee <chenyulee@google.com>}, \
Zifeng Wang$^{2}$, \
Long T. Le$^{2}$, \
Lesly Miculicich$^{2}$,  \\
\bf
Nanyun Peng$^{3}$, \
Chen-Yu Lee$^{2}$, \
Tomas Pfister$^{2}$ \\ 
$^1$University of Southern California, \
$^2$Google Cloud AI Research, \\
$^3$University of California, Los Angeles \\
}

\input{macro}

\begin{document}
\maketitle

\input{00-abstract}
\input{01-intro}

\input{02-prelim_new}

\input{03-method}

\input{04-experimental_setup}
\input{05-results}
\input{10-related}
\input{06-conclusion}

\bibliography{custom}
\bibliographystyle{acl_natbib}

\input{99-appendix}

\end{document}

%% file: macro.tex
\usepackage[flushleft]{threeparttable}
\usepackage{todonotes}
\usepackage{enumitem}
\usepackage{algorithm}
\usepackage[noend]{algorithmic}
\usepackage{booktabs}
\usepackage{subcaption}
\usepackage[nointegrals]{wasysym}
\usepackage{tikz}
\usepackage{tabularx}
\usepackage{tabulary}
\usepackage{listings}
\usepackage{color}
\usepackage{multirow}
\usepackage{colortbl}
\usepackage{framed}
\usepackage{pifont}
\usepackage{dsfont}
\usepackage{xcolor}
\usepackage{amsmath,amsthm}
\usepackage{amssymb}
\usepackage{graphicx}
\usepackage{textcomp}
\usepackage{tcolorbox}
\usepackage{soul}
\usepackage{lipsum}
\usepackage{makecell}
\usepackage{amsfonts}
\usepackage{bbm}
\usepackage{listings}
\setuldepth{Berlin}

\newcommand{\mypar}[1]{\vspace{0.4em}\noindent\textbf{#1 }}

\definecolor{mypurple}{RGB}{121,55,196}
\definecolor{myblue}{RGB}{60,177,245}
\definecolor{myorange}{RGB}{243,206,84}
\definecolor{mygreen}{HTML}{008B72}
\definecolor{myred}{HTML}{FF5733}
\definecolor{light-gray}{gray}{0.9}

\newcommand{\Skip}[1]{{}}

\usepackage{cleveref}
\crefformat{section}{\S~#2#1#3}
\crefformat{subsection}{\S~#2#1#3}
\crefformat{subsubsection}{\S.~#2#1#3}
\crefrangeformat{section}{\S\S#3#1#4 to~#5#2#6}
\crefmultiformat{section}{\S\S#2#1#3}{ and~#2#1#3}{, #2#1#3}{ and~#2#1#3}
\crefmultiformat{subsection}{\S\S#2#1#3}{ and~#2#1#3}{, #2#1#3}{ and~#2#1#3}
\Crefformat{figure}{#2Fig.~#1#3}
\Crefformat{table}{#2Tab.~#1#3}
\Crefformat{appendix}{Appx.~\S#2#1#3}
\crefformat{algorithm}{Alg.~#2#1#3}
\Crefformat{equation}{Eq.~#2#1#3}

\newcommand{\mymodel}{CaLM}

%% file: 00-abstract.tex
\begin{abstract}
Grounded generation aims to equip language models (LMs) with the ability to produce more credible and accountable responses by accurately citing verifiable sources. However, existing methods, by either feeding LMs with raw or preprocessed materials, remain prone to errors. To address this, we introduce \mymodel{}, a novel verification framework. \mymodel{} leverages the insight that a robust grounded response should be consistent with information derived solely from its cited sources.  Our framework empowers smaller LMs, which rely less on parametric memory and excel at processing relevant information given a query, to validate the output of larger LMs.  Larger LM responses that closely align with the smaller LMs' output, which relies exclusively on cited documents, are verified.  Responses showing discrepancies are iteratively refined through a feedback loop.  Experiments on three open-domain question-answering datasets demonstrate significant performance gains of 1.5\% to 7\% absolute average without any required model fine-tuning.

\end{abstract}

%% file: 01-intro.tex
\section{Introduction}
\label{sec:intro}


\begin{figure}[t!]
\centering 
\includegraphics[width=0.99\linewidth]{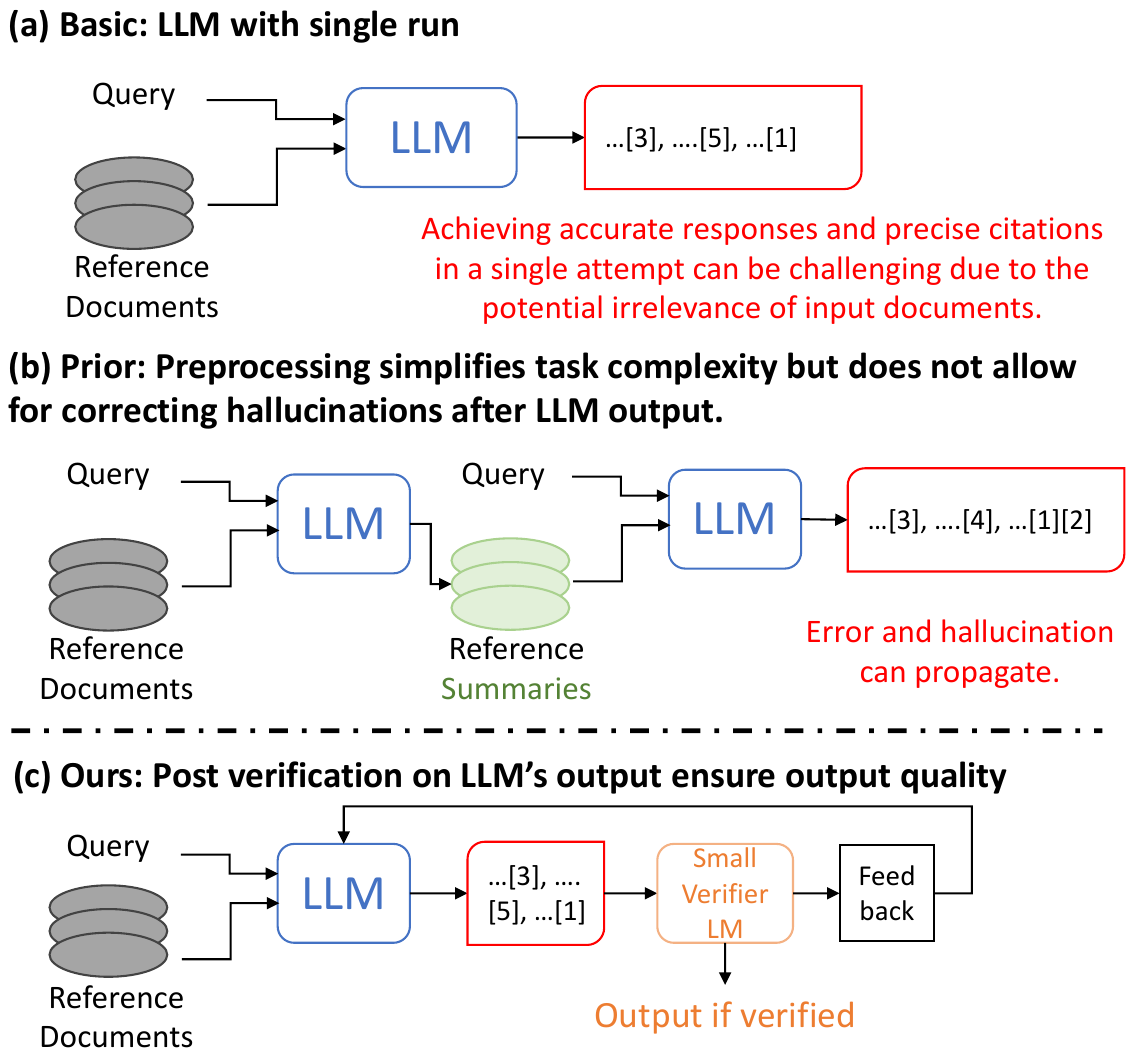} 
\vspace{-5pt}
\caption{
Comparison between different categories of existing inference methods for grounded generation. (a) LLM with single-run can hallucinate easily due to the high complexity of the task. (b) Preprocessing methods reduce task complexity but the hallucination issues can propagate from preprocessing steps. (c)
We propose using verification and rectification to ensure LLMs generate outputs with complete citations and accurate answers, maintaining quality.}
\vspace{-8pt}
\label{fig:teaser} 
\end{figure}

Large Language Models (LLMs) are increasingly popular tools for information seeking. A recent trend emphasizes integrating citations from verifiable sources to boost LLM credibility and enable user verification. This aims to reduce hallucinations and increase accountability~\cite{gao-etal-2023-enabling, DBLP:journals/corr/abs-2307-02185, DBLP:conf/emnlp/LiuZL23}.
To achieve this, LLMs must not only identify relevant documents within vast retrieved collections but also accurately ground their responses in these sources and effectively generate their responses. This significantly increases the complexity of LLM operations~\cite{gao-etal-2023-enabling}.

A standard approach to achieve this grounded generation is by retrieval-augmented generation with instructions to guide LLMs to generate responses along with their corresponding sources in one single LLM inference run (see \Cref{fig:teaser} (a)). More recently, more sophisticated approaches utilize LLMs to first summarize relevant documents \cite{gao-etal-2023-enabling} or use key information extraction and algorithms that explore different relevant document combinations by asking LLMs to enrich the original input query with additional information \cite{LLatrieval2023Li} (see \Cref{fig:teaser} (b)).


However, both single-run (\Cref{fig:teaser} (a)) and preprocessing (\Cref{fig:teaser} (b)) strategies face challenges for accurate generation and citation. Single-run approaches require LLMs to process the input query and a potential large volume of retrieved documents in one forward pass, which can strain their capabilities. Preprocessing approaches, while more focused, risk error propagation or loss of information. Additionally, both strategies limit the LLM's ability to iterate, refine, and verify responses, impacting citation accuracy and answer correctness.

In contrast to single-run and preprocessing strategies, we propose a novel post-verification approach that enables LLMs to fact-check and ground their responses. Our design leverages the complementary strengths of larger and smaller LMs. We observe that larger LMs excel at identifying relevant information within a vast corpus but can rely excessively on internal parametric memory during generation. Smaller LMs, however, are adept at processing retrieved relevant information but less capable of identifying it from large collections (see \Cref{sec:explore_LM} for details).

Building on these observations, we propose \mymodel{} (\textbf{C}ontr\textbf{a}sting \textbf{L}arge and s\textbf{M}all language models to verify grounded generation). \mymodel{} validates the large LLM's response by cross-referencing it with output from a smaller LM. The smaller LM scrutinizes the cited documents to confirm the large LLM's citation accuracy. If the responses align, the large LLM's answer is verified.  If not, \mymodel{} extracts useful statements and evidence from the large LLM's response and seeks additional supporting information to improve the query response. Importantly, \mymodel{} requires no model fine-tuning, allowing smaller LMs to significantly enhance the grounded generation capabilities of large LMs. \Cref{fig:overview} illustrates this process.

We conduct experiments on three open-domain question answering datasets (QAMPARI, ASQA, and ELI5),  which require consulting multiple sources for comprehensive answers. Our method demonstrates significant improvements in both answer accuracy and citation quality, outperforming state-of-the-art methods by an average of 1.5\% to 7\%. Crucially, our method remains robust even in challenging scenarios with less powerful retrieval systems, while other baselines often struggle.

%% file: 02-prelim_new.tex
\section{Problem Statement}
\label{sec:preliminary}

\mypar{Task Setup.}
We cope with the problem of grounded generation~\cite{gao-etal-2023-enabling}.
Given a query $q$ and a corpus of trustworthy text passages $\mathcal{D}$, the model needs to generate an answer response $\mathcal{A}$, which consists of $n$ statements $s_1, s_2, ... s_n$, based on the knowledge in $\mathcal{D}$.
Each statement $s_i$ cites a list of passages $\mathcal{C}_i=\{c_i^1, c_i^2, ...\}, \forall c_i^j \in \mathcal{D}$. 
The collective sets $\mathcal{C}_i$ for $i=1,2,\ldots,n$ constitute the grounded evidence $\mathcal{G}$, from which $\mathcal{A}$ is derived.
Our goal is to jointly optimize the usefulness of $\mathcal{A}$ to $q$, the preciseness of $\mathcal{C}_i$ for statement $s_i$, and the integrity of $\mathcal{G}$ to adequately support $\mathcal{A}$.

\mypar{Evaluation of Response.}
The task involves measuring three dimensions of system responses, following the setup from \citet{gao-etal-2023-enabling}.
\begin{itemize}[noitemsep,nolistsep,leftmargin=*]
  \item \textbf{Fluency}: Determining  whether the model's generated text $\mathcal{A}$ is fluent and coherent.
  \item \textbf{Correctness}: Assessing if $\mathcal{A}$ is accurate and covers \textbf{all relevant aspects} of query $q$.
  \item \textbf{Citation Quality}: Evaluating whether cited passages directly support the answer and avoids irrelevant citations. This is achieved by evaluating both \textbf{citation recall} and \textbf{citation precision}. \citet{gao-etal-2023-enabling} propose measuring citation quality by averaging scores for each statement $s_i$. Citation recall ensures there is at least one supporting citation $c_i^j$ for $s_i$. Citation precision measures whether all the citations are ``relevant''. Specifically, a citation $c_i^j$ is considered ``irrelevant'' to statement $s_i$ if $c_i^j$ cannot support $s_i$, and removing $c_i^j$ from $\mathcal{C}_i$ would not impact the overall support for $s_i$ from the remaining citations.
\end{itemize}

\section{Automated Verification for Grounded Generation}
\label{sec:verification_intro}

Although LLMs have demonstrated proficiency in a wide range of tasks, they remain susceptible to generating hallucinations~\cite{DBLP:journals/corr/abs-2311-05232, DBLP:journals/corr/abs-2309-01219}. These hallucinations could occur in both answers and citations within grounded generations due to the high complexity of the entire working pipeline, which includes noise from the retrievers and the limited ability of LLMs to handle long contexts~\cite{liu-etal:2023:tacl}. This issue underscores the critical need for verification mechanisms to ensure the quality of the generated output and to leverage the interplay between verification and generation systems to improve the final output $\mathcal{A}$.

In this section, we first analyzes key factors to verify grounded generation (\Cref{sec:verification_analysis}). Subsequently, we introduce an automated and unsupervised verification method for grounded generation using a small LM as a verifier and contrasting results from large and small LMs to verify large LMs' response (\Cref{sec:auto_verification} \& \ref{sec:explore_LM}).

\subsection{Key Factors for Automated Verification}
\label{sec:verification_analysis}
Automated verification, unlike the task evaluation in \Cref{sec:preliminary}, operates without a ground-truth reference and should be efficient for real-time system feedback.
Here, our focus lies on assessing answer correctness and citation quality.  
To evaluate the \textbf{correctness} of a generated grounded response, we must ensure that generated responses ($\mathcal{A}$) faithfully leverage information from the knowledge base $\mathcal{D}$, avoiding hallucinations or model biases.
Additionally, \textit{correct reasoning} in deriving the answer is also crucial. 
For the automatic evaluation of \textbf{citation quality}, a trained Natural Language Inference (NLI) model can assess each citation and statement pair iteratively to measure the citation's fidelity~\cite{gao-etal-2023-enabling}. However, this process can be computationally expensive for lengthy generated answers with numerous citations. Efficient automated verification for grounded generation must consider these factors.


\subsection{Contrasting Large and Small LMs for Automated Verification}
\label{sec:auto_verification}
We propose a verification method using a smaller LM to assess the quality of a larger LM's grounded generation.
The small LM receives only the large LM's \underline{\textit{cited documents ($\mathcal{G}$)}} as input to answer the same query $q$. Consistency between their responses indicates the quality of $\mathcal{G}$ and the grounded generation from the large LM.

Our design exploits the inherent characteristics of smaller LMs.
We posit that a robust $\mathcal{G}$ should enable even small LMs to deduce the correct answer.
Notably, small LMs, having fewer parameters, are demonstrably more receptive to integrating external knowledge~\cite{DBLP:journals/corr/abs-2305-13300}. 
Reaching consistent results from both LMs indicate high answer fidelity.~\footnote{In this paper, we differentiate LMs by size, labeling them as large or small. However, a more accurate categorization would be strong versus weak LMs, reflecting their varying performance levels across different LM families.}
Leveraging different LMs as support also \textit{reduce the reasoning error risks} as the different LMs exhibit diverse strengths and reasoning mechanisms~\cite{DBLP:conf/acl/Jiang0L23}. 
These characteristics of small LMs makes our design effective for verifying the answer correctness.

Furthermore, as will detailed in \Cref{sec:explore_LM}, smaller LMs are more sensitive to the relevance of input evidence.
Irrelevant documents in the evidence set $\mathcal{G}$ can easily mislead small LMs, while missing crucial citations hinder their ability to reach the correct answer independently, due to their limited parametric knowledge.
This sensitivity allows us to utilize small LMs for assessing the quality of $\mathcal{G}$.

\begin{figure*}[ht!]
\centering 
\includegraphics[width=0.95\textwidth]{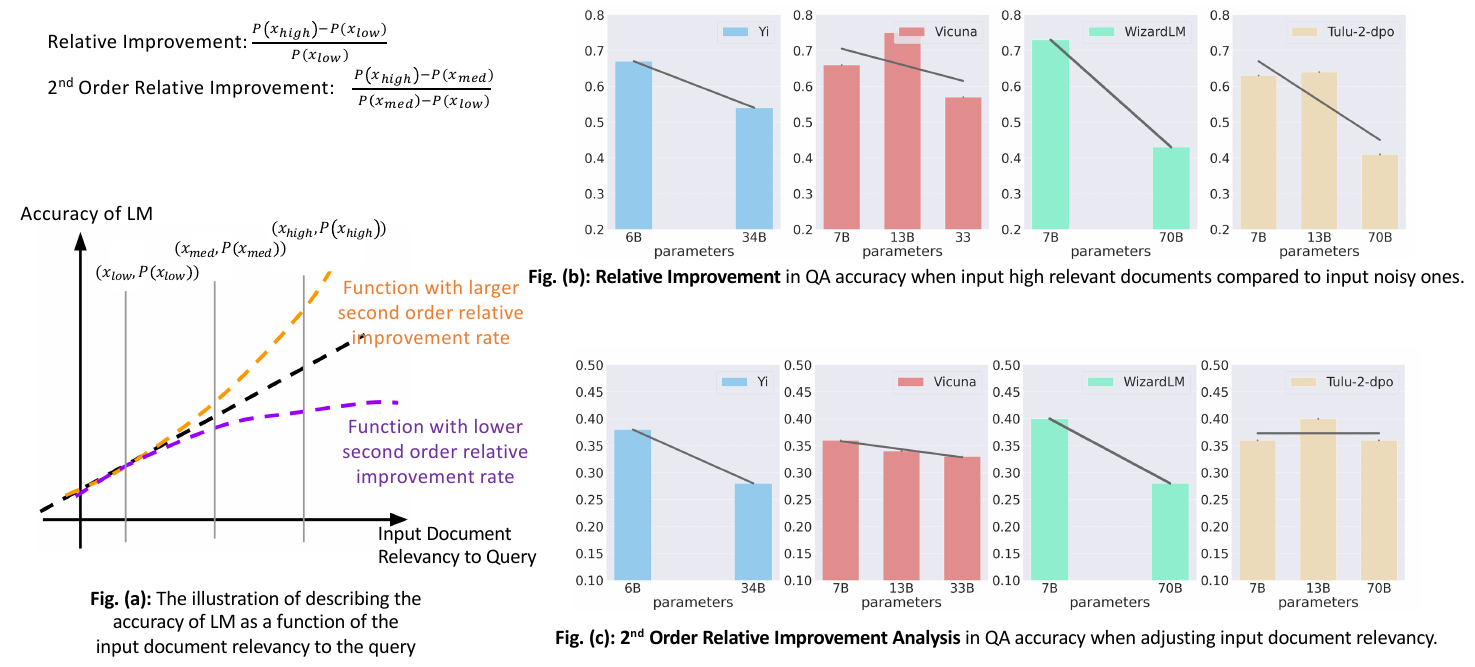} 
\caption{
The studies of the performance of an LM as a function of the input document's relevance score using the ASQA dataset.
We show that, \textit{within the same LM family, smaller LMs demonstrate higher sensitivity to the relevance of the input document, when anchored to the largest model in the family}.
\textbf{\textit{(a)}} The illustration of the function. This function is a monotonic increase function as the accuracy always increase when input document's relevance score increase. Hence, studying the second order relative improvements can help us know the \textit{incremental performance gain} for the LM when the input document's relevance keep increasing.
\textbf{\textit{(b)}} The result of relative improvement.
\textbf{\textit{(c)}} The result of the second order relative improvement analysis. 
From (b)(c), we can observe that smaller models tend to exhibit greater relative improvements and achieve larger incremental performance gains compared to their larger counterparts.
}
\vspace{-6pt}
\label{fig:prelim} 
\end{figure*}

\subsection{Analyzing Model Size Impact on LMs' Sensitivity to Input Document Relevance}
\label{sec:explore_LM}
Our automated verification method exploits the high sensitivity of small LMs. This section empirically demonstrates that, within the same model family, smaller LMs exhibit greater sensitivity to the relevance of input documents to a given query $q$ compared to their larger counterparts.

We investigate the sensitivity of LMs to the relevance of input documents by modeling the performance of an LM as a function of the input document's relevance. 
To mitigate the effects of the divergent abilities for different LMs to follow instructions, we examine the \textit{relative} performance improvement of the model response as the relevance of the input document increases. 
The larger this value is, the more sensitive the LM is.

Yet, the complexity of this function is beyond simple linearity. As depicted in \Cref{fig:prelim}(a), this function typically exhibits a monotonic increase and hence we intend to additionally study second-order improvements gain analysis to further study the curvature of the function.
This analysis use three anchoring points $x_{low}, x_{med}, x_{high}$, and the corresponding performance $P(x_{low}), P(x_{med}), P(x_{high})$ to study the much incremental performance gain the LM can get when the input document's relevance keep increasing. Specifically, we use the ratio $\tfrac{P(x_{high})-P(x_{med})}{P(x_{med})-P(x_{low})}$ to represent the incremental performance gain.
A higher ratio indicates the LM is more sensitive to input document's relevance, as illustrated by the orange line in \Cref{fig:prelim}(a).

We conduct empirical studies using a retrieval-augmented generation setting on the ASQA dataset~\cite{DBLP:conf/emnlp/StelmakhLDC22}. Each instance in the dataset contains multiple answers and requires reading multiple documents to make correct prediction. 
In this experiment, a LM utilizes five input documents $\mathbf{d}$ to answer queries $q$, with model performance evaluated based on answer accuracy, and the relevance of the input documents $\mathbf{d}$ is assessed by their average recall rate in containing answers.
We prepare three different set of $\mathbf{d}$ with approximately 27\%, 56\%, and 78\% recall rates, respectively, to represents $x_{low}, x_{med}, x_{high}$ in \Cref{fig:prelim}(a).

\Cref{fig:prelim}(b) presents the results of relative performance improvement, while \Cref{fig:prelim}(c) shows the second-order analysis. Our study encompasses LM families such as Yi-LM, Vicuna~\cite{DBLP:journals/corr/abs-2306-05685}, WizardLM~\cite{DBLP:journals/corr/abs-2304-12244}, and Tulu-2-dpo~\cite{DBLP:journals/corr/abs-2311-10702}. 
The findings indicate that within the same LM family, compared to the largest model, smaller LMs usually achieve higher relative improvement and receive a greater incremental performance gain, suggesting \textit{their higher sensitivity to the relevance of input documents to the query $q$}.
On the other hand, smaller LMs are more prone to erroneous results and suboptimal performance with irrelevant documents, which motivates our automated verification method design.

%% file: 03-method.tex
\begin{figure*}[ht!]
\centering 
\includegraphics[width=0.95\textwidth]{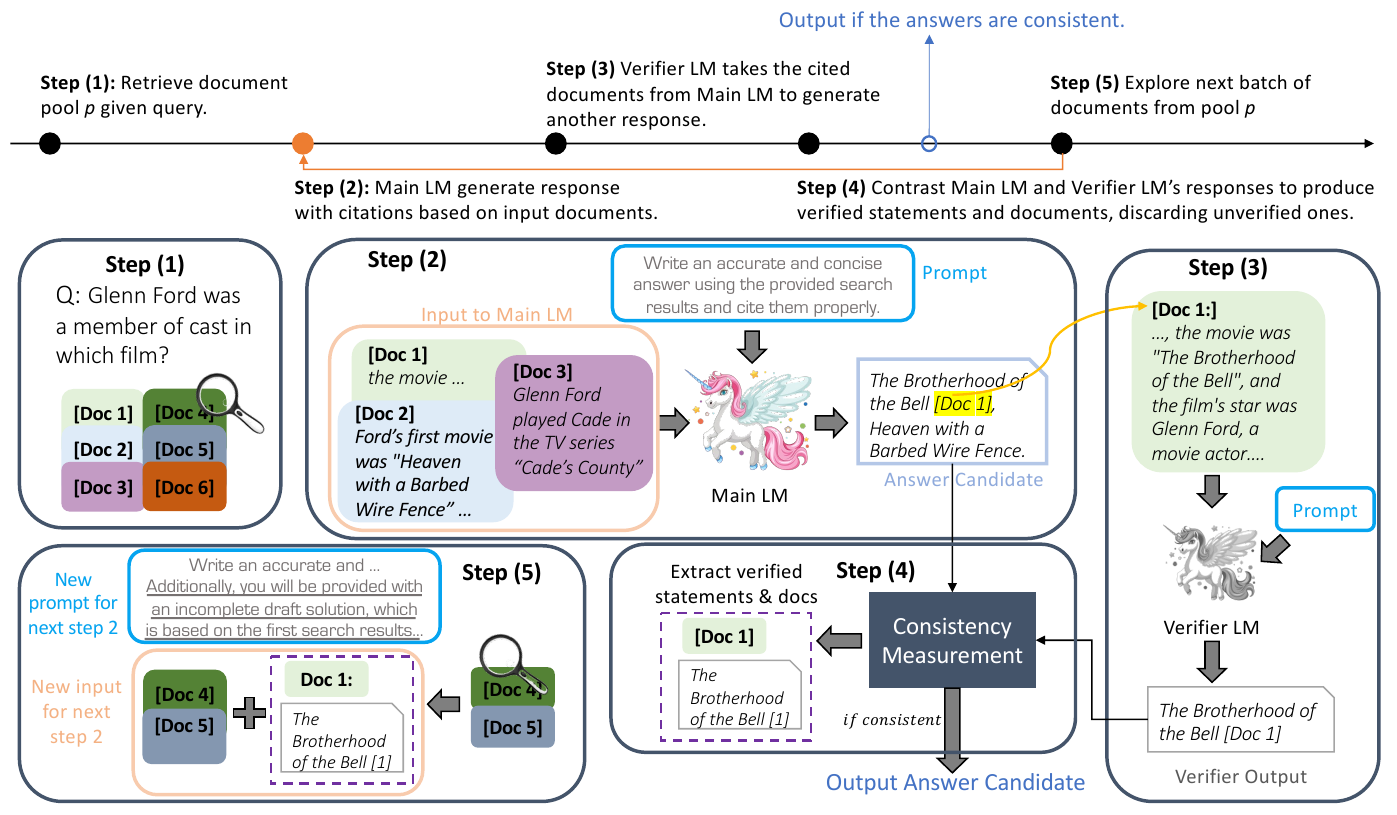} 
\caption{Overview of \mymodel{}: \textit{\textbf{Top:}} The flow diagram of our method. \textit{\textbf{Bottom:}} A detailed depiction of each step's operation. The algorithm starts with a retriever extract a relevant document pool $p$ for the input query (Step (1)).
Then, the main language model (LM) takes the first batch of documents and employs retrieval-augmented generation to produce an answer candidate, which cites relevant supporting documents (Step (2)). 
Subsequently, this candidate is validated by contrasting it with the verifier output from the verifier LM (Steps (3) \& (4)). 
Our verifier LM evaluates citation quality by accessing only the documents cited by the main LM's response, rather than the same input documents. For responses with sufficient consistency, we accept the answer candidate directly. 
If inconsistent, we break down the answer candidate into individual statements, retaining only those corroborated by similar arguments in the verifier output for further correction in next iteration (Step (5)).}
\vspace{-6pt}
\label{fig:overview} 
\end{figure*}

\section{\mymodel{} Framework}
\label{sec:method}
Building on the automated verification design from \Cref{sec:auto_verification}, we introduce \mymodel{}, an inference framework that leverages the synergy between verification and generation systems to enhance the system's final grounded generation.

\Cref{fig:overview} depicts the iterative five-step algorithm of our \mymodel{} framework. Steps three and four correspond to the automated verification method detailed in \Cref{sec:auto_verification}. The remaining steps involve the large LM making predictions. 

We now present a detailed breakdown of each step below. We differentiate between the ``main LM,'' whose responses are verified, and the ``verifier LM,'' an auxiliary LM assisting the verification process. As described in \Cref{sec:auto_verification}, we recommend using a larger LM as the main LM and a smaller LM for verification to achieve optimal performance.

\mypar{Step (1) Context Retrieval.} \mymodel{} starts by selecting a \textit{ranked pool} of trustworthy passages $p$ for a given query $q$ using a retriever. 

\mypar{Step (2) Main LM Generation.} We select the top-$k$ documents from the reference pool $p$ and feed them into the main LM. This value of $k$ is a hyperparameter constrained by the maximum input capacity of the main LM. The LM then analyzes these $k$ passages to generate an answer candidate $\mathcal{A}$ for the query $q$. Our findings in \Cref{sec:explore_LM} suggest employing a large-scale LM at this stage. This is because larger LMs exhibit greater robustness in accurately identifying useful information and filtering out noise within the retrieved passages.

\mypar{Step (3) Verifier LM Generation.} 
Building upon the automated verification design outlined in \Cref{sec:auto_verification}, this step leverages a grounded evidence set $\mathcal{G}$ extracted from the main LM's answer candidate $\mathcal{A}$. A smaller, dedicated verification LM then re-attempt the grounded generation process and obtain verifier output $\mathcal{A'}$ for query $q$ by solely providing the model with the evidence set, $\mathcal{G}$, rather than the full top-$k$ documents.

\mypar{Step (4) Contrasting Answer Candidate and Verifier Output:} 
A strong evidence $\mathcal{G}$ should enable small LMs to deduce correct answers reliably. Our goal in this step is to verify consistency~\cite{DBLP:conf/iclr/0002WSLCNCZ23} of $\mathcal{A'}$ against the answer candidate $\mathcal{A}$.

If the comparison shows enough consistency, we accept the answer $\mathcal{A}$ and stop the iterative process. 
Otherwise, we dismiss the inconsistent segments from the answer and citations, and continue with the next iteration.
More technically, we extract $\bar{\mathcal{A}} = \mathcal{A} \cap \mathcal{A'}$ and the corresponding $\bar{\mathcal{G}}$ within $\bar{\mathcal{A}}$ for Step (5) usage.

The realization of consistency measurement is done via calculating ROUGE-2 score between the $\mathcal{A'}$ and $\mathcal{A}$. If the ROUGE-2 score exceeds a threshold $\theta$, the answer candidate is considered acceptable. This threshold can be tuned using a small development set. Empirically, by our observation, setting $\theta=0.2$ to $0.5$ yields satisfactory results by small set of qualitative examination.

\begin{table*}[t!]
\centering
\small
\setlength{\tabcolsep}{3.5pt}
\renewcommand{\arraystretch}{1}
\begin{tabular}{l|p{13.8cm}}
    \toprule
    \textbf{Dataset} & \textbf{Example} \\
    \midrule
    \multirow{2}{*}{ASQA} & \textbf{Question:} Who sings don't tell me what to do?\\
    & \textbf{Reference Answer:} Marty Stuart recorded the song Don't Tell Me What to Do, recorded as I'll Love You Forever (If I Want To) in 1988. Pam Tillis sang Don't Tell Me What To Do in 1990, and in 1993, the Baby Animals recorded the song.\\
    \midrule
    \multirow{2}{*}{QAMPARI} & \textbf{Question:}  Which movie did John Carpenter direct for which he also composed the music?\\
    & \textbf{Reference Answer:} Vampires, In the Mouth of Madness, Assault on Precinct 13, Dark Star, Big Trouble in Little China, They Live, Halloween, Escape from New York, Prince of Darkness, Ghosts of Mars, The Fog, Chevil, Village of the Damned\\
    \midrule
    \multirow{2}{*}{ELI5} & \textbf{Question:} Why are fruit in Chinese supermarkets so much bigger than western chains like loblaws or food basics?\\
    & \textbf{Reference Answer:} There are a lot less restrictions on using chemicals in agriculture here, so some farmers use 'growth accelerators' to make their fruit (and other products) bigger. Last May, overuse of the chemicals resulted in a spate of [exploding watermelons]\\
    \bottomrule
\end{tabular}
\caption{
Illustrative examples from each experiment's dataset.
}
\vspace{-6pt}
\label{table:dataset_example}
\end{table*}

\mypar{Step (5) Input Preparation for Next Iteration:} 
In this final step, we prepare input for the next iteration, including input reference document lists and the draft for correction.
The new input reference document lists is initialized by $\bar{\mathcal{G}}$ and is supplemented more passages from the next batch of passages from pool $p$ until complete the budget $k$.
This process boosts the likelihood of finding relevant documents while maintaining useful documents that have been verified.
Then, we create a new prompt for large LM focusing on leveraging the new input reference document lists to correct the incomplete answer response $\bar{\mathcal{A}}$.

The full prompt we use for each step are detailed in \Cref{appx:prompt}. In practice, we should set a maximum iteration $T$ to halt the whole process to prevent the verification condition cannot be satisfied. If this maximum iteration is reached, we will output the last answer candidate we get from the main LM.

%% file: 04-experimental_setup.tex
\section{Experimental Setup}
\label{sec:experimental_setup}
We consider three factoid question-answering (QA) datasets. For each dataset, we present an illustrative example in \Cref{table:dataset_example} for better understanding.

We prepare the trustworthy text passages $\mathcal{D}$ for each dataset accordingly following \cite{gao-etal-2023-enabling}.
Each entry in $\mathcal{D}$ is a 100-word passage following previous works on open-domain QA~\cite{DPR2020, DBLP:conf/naacl/PetroniPFLYCTJK21, liu-etal:2023:tacl}.

\subsection{The ASQA dataset}
\mypar{Basic Introduction.}
ASQA \cite{DBLP:conf/emnlp/StelmakhLDC22} is a long-form generation QA dataset derived from the AmbigQA dataset~\cite{DBLP:conf/emnlp/MinMHZ20}. It comprises questions characterized by their ambiguity, necessitating multiple short answers to address various aspects. 
Each entry in the dataset is accompanied by a comprehensive long-form answer that covers all the corresponding short answers. 948 samples are tested for our experiment.

\mypar{Experimental Setting}
Since most questions can be answered by Wikipedia, prior works usually use 2018-12-20 Wikipedia snapshot as $\mathcal{D}$. 

For retriever utilization, we examine the application of both DPR~\cite{DPR2020} and GTR-large~\cite{DBLP:conf/emnlp/Ni0LDAMZLHCY22}. DPR introduces marginally more complex challenges for the LLM, achieving a recall rate of 51.5\%, whereas GTR achieves 56.8\% when considering the top-$5$ retrieved documents.

\mypar{Evaluation}
\begin{itemize}[noitemsep,nolistsep,leftmargin=*]
  \item \textbf{Fluency}: We use \textbf{MAUVE} score~\cite{DBLP:conf/nips/PillutlaSZTWCH21} to evaluate the corpus-wise similarity of the machine generated text and the long answers generated by systems.
  \item \textbf{Correctness}: We follow \cite{DBLP:conf/emnlp/StelmakhLDC22} to calculate the exact matching recall (\textbf{EM recall}) of the presents of correct short answers.
  \item \textbf{Citation quality}: As mentioned in \Cref{sec:preliminary}, we calculate \textbf{citation recall} and \textbf{citation precision} using the scripts provided by \cite{gao-etal-2023-enabling}.
\end{itemize}

\subsection{The QAMPARI dataset}
\mypar{Basic Introduction.}
QAMPARI~\cite{DBLP:journals/corr/abs-2205-12665} is created from Wikipedia, pairing questions with multiple answers derived from its knowledge graph and tables. 
These answers, comprised of entities, describe simple relationships to the entities in the query $q$. 
As a result, this dataset focuses on testing systems' abilities on entity identification within questions and accurately pinpointing the relevant entities.
We use the same 1000 testing samples used in \cite{gao-etal-2023-enabling} for experiments.

\mypar{Experimental Setting}
Similar to the ASQA case, we employ the Wikipedia snapshot from 2018-12-20, as our  $\mathcal{D}$. For retrieval, we again use both DPR and GTR-large, achieving recall rates of approximately 17.6\% and 31.6\%, respectively, for the top five retrieved documents for each query.

\mypar{Evaluation Metrics}
In QAMPARI, we only considering the correctness and the citation quality since the output of the dataset is a list of entities. 
\begin{itemize}[noitemsep,nolistsep,leftmargin=*]
  \item \textbf{Correctness}: We follow \cite{DBLP:conf/emnlp/StelmakhLDC22} to to calculate the entity \textbf{precision} and recall of the model prediction using \textit{exact string match}. When calculate recall, the evaluation considers recall to be 100\% if the prediction includes at least 5 correct answer, denoted as \textbf{recall-5}.
  \item \textbf{Citation quality}: We use the same way as the ASQA dataset to evaluate the citation quality.
\end{itemize}


\subsection{The ELI5 dataset}
\mypar{Basic Introduction.}
The ELI5 dataset, introduced by \cite{DBLP:conf/acl/FanJPGWA19}, primarily features ``How,'' ``Why,'' and `What'' questions. It tests a system's ability to summarize complex information into clear and insightful answers.
We use the 1000 samples used in \cite{gao-etal-2023-enabling} for our experiment.

\mypar{Experimental Setting}
Unlike ASQA and QAMPARI, the ELI5 dataset covers diverse topics and hence, documents in Sphere corpus are treated as $\mathcal{D}$~\cite{DBLP:journals/corr/abs-2112-09924}. Given the large size of the Sphere corpus, BM25 is used for efficient retrieval.

\mypar{Evaluation Metrics}
\begin{itemize}[noitemsep,nolistsep,leftmargin=*]
  \item \textbf{Correctness}: ELI5 dataset does not provide short entity answers. We follow \cite{gao-etal-2023-enabling} to calculate \textbf{claim recall} for correctness. For each reference answer in the dataset, three ``sub-claims'' are first extracted, and we will test whether the machine's answer $\mathcal{A}$ can entail these sub-claims using a TRUE NLI model~\cite{honovich-etal-2022-true-evaluating}
  \item \textbf{Fluency} \& \textbf{Citation quality}: We use the metrics as the ASQA dataset for evaluation.
\end{itemize}

\subsection{Compared Methods}
We compare the following methods, all of which we have independently rerun, except Self-RAG. For our own rerun results, the reported results are the average of three random runs.
\begin{enumerate}[noitemsep,nolistsep,leftmargin=*]
    \item \textbf{In-Context Learning (ICLCite)}: LLMs are invoked once for grounded generation through instruction-based in-context learning. Following \cite{gao-etal-2023-enabling}, we provide five documents to the LLM. They suggest that increasing the number of input document lists does not significantly improve performance when using GPT-3.5-turbo.
    \item \textbf{Summary then In-Context Learning (Summ + ICLCite)}~\cite{gao-etal-2023-enabling}: This method follows the preprocessing paradigm shown in \Cref{fig:teaser}. Initially, LLMs generate summaries for each document based on the query $q$. Then, these summaries are fed into the LLM to execute ICLCite. In our experiments, we generate summaries for the top-9 documents retrieved for each instance.
    \item \textbf{Snippet then In-Context Learning (Snippet + ICLCite)}~\cite{gao-etal-2023-enabling}: Similar to Summ + ICLCite, but LLM are guided to generate extractive summaries during preprocessing steps.
    \item \textbf{In-Context Learning with Self-Consistency (ICLCite + USC)}: This post-processing method that employs ICLCite to initially generate various output samples. Then, it applies universal self-consistency \citet{chen2023universal} to obtain the results. For a fair comparison with other baselines, like Summ+ICLCite, we first generate 9 samples and then use a LLM to determine the most consistent result among them.
    \item \textbf{Self-RAG}~\cite{asai2023self}: This method finetunes LMs to generate special tokens to trigger additional fact checks and retrieval. Since this method requires model finetuning, we only report results on ASQA dataset only.
     \item \textbf{\mymodel{}}: We use verification and an iterative refinement design to ensure the output quality. We follow the setting of ICLCite to set the our main LM's reading budget $k=5$. We set the consistency threshold $\theta = 0.25$ for the ELI5 dataset and $\theta = 0.5$ for the ASQA dataset. This $\theta$ is decided by manual qualitative examination on a small set of development data.
     Besides, we set the maximum iteration to be 4 for budget concern, and use the 13B version of tulu-2-dpo~\cite{DBLP:journals/corr/abs-2311-10702} as the verifier LM. 
\end{enumerate}

\begin{table*}[t!]
\centering
\small
\resizebox{1.0\textwidth}{!}{
\setlength{\tabcolsep}{3.5pt}
\renewcommand{\arraystretch}{1}
\begin{tabular}{lc|cc|ccccc|ccccc}
    \toprule
    \multirow{3}{*}{\textbf{Method}}  & \multirow{3}{*}{\textbf{\shortstack{Method \\ Type}}} & \multirow{3}{*}{\textbf{\shortstack{\# Main \\ LM \\ Call}}}  & \multirow{3}{*}{\textbf{\shortstack{\# Verifier \\ LM  \\ Call}}}  & \multicolumn{5}{c|}{DPR as retriever} & \multicolumn{5}{c}{GTR as retriever} \\
     \cmidrule{5-14}
    & & & & \textbf{Fluency} & \textbf{Correct.} & \multicolumn{2}{c}{\textbf{Citation}}  & \multirow{2}{*}{\textbf{Average}} & \textbf{Fluency} & \textbf{Correct.} & \multicolumn{2}{c}{\textbf{Citation}} & \multirow{2}{*}{\textbf{Average}}\\
    & & & & mauve & EM Rec. & Prec. & Rec. & & mauve & EM Rec. & Prec. & Rec. &\\
    \midrule
    \multicolumn{13}{c}{\textbf{GPT-3.5-Turbo-1106 as Main LM}} \\
    \midrule
    ICLCite~\cite{gao-etal-2023-enabling} & Single Run & 1 & - 
    & 74.73 & 39.32 & \textbf{67.36} & \ul{69.48} & \ul{62.72} 
    & 71.85 & 41.92 & \textbf{73.14} & \ul{77.90} & \ul{66.20}
    \\
    \cline{2-2}
    Summ + ICLCite~\cite{gao-etal-2023-enabling} & \multirow{2}{*}{Preprocess}& 10 & - 
    & 48.95 & 29.30 & 60.14 & 54.52 & 48.23 {\scriptsize \color{myred}{(-14.49)}}
    & 68.01 & 41.11 & 66.04 & 74.43 & 62.40 {\scriptsize \color{myred}{(-3.8)}}\\
    Snippet + ICLCite~\cite{gao-etal-2023-enabling} & & 10 & - 
    & 48.56 & 29.48 & 59.52 & 53.84 & 47.85  {\scriptsize \color{myred}{(-14.87)}}
    & 68.84 & 39.89 & 62.05 & 71.06 & 60.46 {\scriptsize \color{myred}{(-5.74)}}\\
    \cline{2-2}
    ICLCite + USC~\cite{chen2023universal}$^*$ & \multirow{2}{*}{Postprocess} & 10 & - 
    & \ul{77.50} & \ul{40.71} & 61.20 & 64.07 & 60.87 {\scriptsize \color{myred}{(-1.85)}}
    & \ul{77.31} & \ul{42.75} & 67.08 & 71.64 & 64.69  {\scriptsize \color{myred}{(-1.51)}}\\
    
    \mymodel{} (ours) & & $\leq$ 4 &  $\leq$ 3
    & \textbf{81.35} & \textbf{43.56} & \ul{66.00} & \textbf{69.95} & \textbf{64.71} {\scriptsize \color{mygreen}{(+1.99)}}
    & \textbf{83.98} & \textbf{45.01} & \ul{72.59} & \textbf{78.03} & \textbf{68.98} {\scriptsize \color{mygreen}{(+2.78)}} \\
    
    \midrule
    \multicolumn{13}{c}{\textbf{text-unicorn as Main LM}} \\
    \midrule
    ICLCite~\cite{gao-etal-2023-enabling} & Single Run & 1 & - 
    & 62.01 & 37.09 & \ul{62.42} & \ul{60.35} & \ul{55.46} 
    & 63.25 & 39.83 & \ul{69.39} & 67.98 & 60.11\\
    \cline{2-2}
    Summ + ICLCite~\cite{gao-etal-2023-enabling}  & \multirow{2}{*}{Preprocess}& 10 & - 
    & \ul{63.21} & \ul{38.67} & 52.41 & 59.45 & 53.43 {\scriptsize \color{myred}{(-2.03)}}
    & \ul{75.68} & \ul{42.65} & 61.18 & \ul{68.91} & \ul{62.11} {\scriptsize \color{mygreen}{(+2.00)}}\\
    Snippet + ICLCite~\cite{gao-etal-2023-enabling} & & 10 & - 
    & 59.03 & 37.69 & 54.62 & 59.44 & 52.69 {\scriptsize \color{myred}{(-2.77)}}
    & 72.50 & 40.97 & 60.88 & 68.02 & 60.59  {\scriptsize \color{mygreen}{(+0.48)}}\\
    \cline{2-2}
    ICLCite + USC~\cite{chen2023universal} & \multirow{2}{*}{Postprocess} & 10 & - 
    & 57.92 & 37.16 & 62.05 & 60.00 & 54.28  {\scriptsize \color{myred}{(-1.18)}}
    & 63.27 & 40.75 & 68.90 & 67.60 & 60.13 {\scriptsize \color{mygreen}{(+0.02)}}\\
    
    \mymodel{} (ours) & & $\leq$ 4 &  $\leq$ 3
    & \textbf{77.18} & \textbf{42.24} & \textbf{63.71} & \textbf{64.99} & \textbf{62.03} {\scriptsize \color{mygreen}{(+6.57)}}
    & \textbf{82.08} & \textbf{44.21} & \textbf{70.55} & \textbf{72.37} & \textbf{67.30} {\scriptsize \color{mygreen}{(+7.19)}}\\

    \midrule
    \multicolumn{13}{c}{\textbf{Finetune Llama-2 Baseline}} \\
    \midrule
    Self-RAG (7B) $^{\dagger}$ & Finetune LM & - 
    & - & - & - & - & - &-
    & 74.3 & 30.0 & 66.9 & 67.8 & 59.8 \\
    Self-RAG (13B) $^{\dagger}$ & Finetune LM & - 
    & - & - & - & - & - & -
    & 71.6 & 31.7 & 70.3 & 71.3 & 61.2 \\
    \bottomrule
\end{tabular}
}
\caption{
The experimental results on ASQA. \mymodel{} achieves an average improvement of over 6\% when using text-unicorn. When using GPT-3.5-Turbo-1106 as the main LM, \mymodel{} is the only method that outperforms the ICLCite baseline while making the fewest total LM API calls. 
 The best results are bold, while the second best are underlined. $^*$USC stands for Universal Self Consistency~\cite{chen2023universal}. $^{\dagger}$We report Self-RAG's numbers using the results from their original paper, where they retrieve up to ten documents per input using Contriever as the retriever~\cite{DBLP:journals/tmlr/IzacardCHRBJG22}.
}
\label{table:main_asqa_result}
\end{table*}

\begin{table*}[ht!]
\centering
\small
\resizebox{1.0\textwidth}{!}{
\setlength{\tabcolsep}{3.5pt}
\renewcommand{\arraystretch}{1.03}
\begin{tabular}{lc|cc|ccccc|ccccc}
    \toprule
    \multirow{3}{*}{\textbf{Method}} & \multirow{3}{*}{\textbf{\shortstack{Method \\ Type}}} & \multirow{3}{*}{\textbf{\shortstack{\# Main \\ LM \\ Call}}}  & \multirow{3}{*}{\textbf{\shortstack{\# Verifier \\ LM  \\ Call}}}  & \multicolumn{5}{c|}{DPR as retriever} & \multicolumn{5}{c}{GTR as retriever} \\
     \cmidrule{5-14}
    & & & & \multicolumn{2}{c}{\textbf{Correctness}} & \multicolumn{2}{c}{\textbf{Citation}}  & \multirow{2}{*}{\textbf{Average}} & \multicolumn{2}{c}{\textbf{Correctness}} & \multicolumn{2}{c}{\textbf{Citation}} & \multirow{2}{*}{\textbf{Average}}\\
    & & & & Prec. & Rec.-5 & Prec. & Rec. & & Prec. & Rec.-5 & Prec. & Rec. &\\
    \midrule
    \multicolumn{13}{c}{\textbf{GPT-3.5-Turbo-1106 as Main LM}} \\
    \midrule
    ICLCite~\cite{gao-etal-2023-enabling} & Single Run & 1 & - 
    & \ul{12.47} & \ul{10.28} & 12.60 & 11.62 & \ul{11.74}
    & 19.23 & \ul{17.32} & 21.75 & 20.77 & 19.77\\
    \cline{2-2}
    Summ + ICLCite~\cite{gao-etal-2023-enabling} & \multirow{2}{*}{Preprocess} & 10 & - 
    & 6.65 & 4.97 & 10.56 & 9.43 & 7.90 {\scriptsize \color{myred}{(-3.84)}}
    & 15.06 & 13.10 & 21.89 & 21.00 & 17.76 {\scriptsize \color{myred}{(-2.01)}}\\
    Snippet + ICLCite~\cite{gao-etal-2023-enabling} & & 10 & - 
    & 11.67 & 7.87 & \textbf{14.14} & \ul{12.64} & 11.58 {\scriptsize \color{myred}{(-0.16)}}
    & \ul{20.48} & 17.18 & \textbf{25.45} & \ul{23.91} & \ul{21.76} {\scriptsize \color{mygreen}{(+1.99)}}\\
    \cline{2-2}
    ICLCite + USC~\cite{chen2023universal}$^*$ & \multirow{2}{*}{Postprocess} & 10 & - 
    & 9.88 & 8.36 & 10.14 & 9.28 & 9.42 {\scriptsize \color{myred}{(-2.02)}}
    & 14.07 & 12.45 & 17.34 & 16.60 & 15.12 {\scriptsize \color{myred}{(-4.65)}}\\
    
    \mymodel{} (ours) & & $\leq$ 4 &  $\leq$ 3
    & \textbf{17.65} & \textbf{13.61} & \ul{13.93} & \textbf{12.99} & \textbf{14.55} {\scriptsize \color{mygreen}{(+2.81)}}
    & \textbf{26.71} & \textbf{18.55} & \ul{25.16} & \textbf{24.39} & \textbf{23.70} {\scriptsize \color{mygreen}{(+3.93)}}\\
    
    \midrule
    \multicolumn{13}{c}{\textbf{text-unicorn as Main LM}} \\
    \midrule
    ICLCite~\cite{gao-etal-2023-enabling} & Single Run & 1 & - 
    & 17.06 & 12.55 & 15.77 & 15.59 & 15.24
    & 26.27 & 21.42 & 25.52 & 25.21 & 24.61 \\
    \cline{2-2}
    Summ + ICLCite~\cite{gao-etal-2023-enabling} & \multirow{2}{*}{Preprocess} & 10 & - 
    & 18.61 & \ul{15.17} & 15.61 & 15.36 & 16.19 {\scriptsize \color{mygreen}{(+0.95}}
    & 27.11 & \textbf{26.14} & 24.61 & 24.98 & 25.71 {\scriptsize \color{mygreen}{(+1.10)}}\\
    Snippet + ICLCite~\cite{gao-etal-2023-enabling} & & 10 & - 
    & \ul{18.81} & 15.11 & \textbf{16.10} & \ul{15.83} & \ul{16.46} {\scriptsize \color{mygreen}{(+1.22)}}
    & \ul{27.26} & \ul{25.07} & \ul{25.96} & \ul{25.55} & \ul{25.96} {\scriptsize \color{mygreen}{(+1.35)}}\\
    \cline{2-2}
    ICLCite + USC~\cite{chen2023universal} & \multirow{2}{*}{Postprocess} & 10 & - 
    & 17.04 & 12.47 & 15.81 & 15.62 & 15.24 {\scriptsize \color{mygreen}{(+0.0)}}
    & 26.34 & 22.10 & 25.48 & 25.19 & 24.78 {\scriptsize \color{mygreen}{(+0.17)}}\\
    
    \mymodel{} (ours) & & $\leq$ 4 &  $\leq$ 3 
    & \textbf{19.62} & \textbf{16.71} & \ul{15.90} & \textbf{15.95} & \textbf{17.05}  {\scriptsize \color{mygreen}{(+1.81)}}
    & \textbf{30.28} & 22.79 & \textbf{28.53} & \textbf{28.34} & \textbf{27.49}  {\scriptsize \color{mygreen}{(+2.88)}}\\
    \bottomrule
\end{tabular}
}
\caption{The experimental results on QAMPARI. Compared to all the preprocess and postprocess baselines, our method obtains the best average performance across different settings and use the fewest LM API calls.}
\vspace{-6pt}
\label{table:main_qampari_result}
\end{table*}

\begin{table}[ht!]
\centering
\small
\resizebox{1.0\columnwidth}{!}{
\setlength{\tabcolsep}{3.5pt}
\renewcommand{\arraystretch}{1.03}
\begin{tabular}{l|ccccc}
    \toprule
    \multirow{2}{*}{\textbf{Method}} & \textbf{Fluency} & \textbf{Correct.} & \multicolumn{2}{c}{\textbf{Citation}} & \multirow{2}{*}{\textbf{Average}}\\
    & mauve & Claim. & Prec. & Rec. & \\
    \midrule
    \multicolumn{6}{c}{\textbf{GPT-3.5-Turbo-1106 as Main LM}} \\
    \midrule
    ICLCite 
    & \ul{25.33} & \ul{12.90} & \ul{44.63} & \ul{49.34} & \ul{33.05} \\
    Summ + ICLCite 
    & 17.86 & 10.67 & 38.81 & 43.26 & 26.40 {\scriptsize \color{myred}{(-6.65)}}\\
    Snippet + ICLCite
    & 25.30 & 12.06 & 35.49 & 41.35 & 28.55 {\scriptsize \color{myred}{(-4.50)}}\\
    ICLCite + USC
    & 25.04 & 12.16 & 34.87 & 39.60 & 27.92 {\scriptsize \color{myred}{(-5.13)}}\\
    \mymodel{} (ours) 
    & \textbf{25.84} & \textbf{12.92} & \textbf{46.55} & \textbf{51.90} & \textbf{34.30} {\scriptsize \color{mygreen}{(+1.25)}}\\

    \midrule
    \multicolumn{6}{c}{\textbf{text-unicorn as Main LM}} \\
    \midrule
    ICLCite 
    & 35.82 & 12.21 & 35.35 & 32.05 & 28.86 \\
    Summ + ICLCite 
    & 34.57 & 11.74 & \ul{35.54} & \ul{35.02} & 29.22 {\scriptsize \color{mygreen}{(+0.36)}}\\
    Snippet + ICLCite 
    & \textbf{47.43} & \textbf{13.51} & 34.37 & 33.00 & 32.08 {\scriptsize \color{mygreen}{(+3.22)}}\\
    ICLCite + USC   
    & 31.42 & \ul{12.29} & 34.90 & 31.66 & 27.57 {\scriptsize \color{myred}{(-1.29)}}\\
    
    \mymodel{} (ours) 
    & \ul{37.06} & 11.95 & \textbf{46.26} & \textbf{43.60} & \textbf{34.72} {\scriptsize \color{mygreen}{(+5.86)}}\\
    \bottomrule
\end{tabular}
}
\caption{The experimental results on ELI5. \mymodel{} obtains the best average performance regardless of the used main LM. The improvements are especially significant in the citation quality.}
\vspace{-6pt}
\label{table:main_eli5_result}
\end{table}

%% file: 05-results.tex
\section{Experimental Results}
\label{sec:results}

\subsection{Main Results}
For the main experiment, we consider two different large LMs as the backbone: GPT-3.5-Turbo-1106 ~\cite{DBLP:conf/nips/Ouyang0JAWMZASR22} and the PaLM-based LLMs~\cite{DBLP:journals/corr/abs-2305-10403}, text-unicorn~\footnote{https://cloud.google.com/vertex-ai/docs/generative-ai/model-reference/text}.

\Cref{table:main_asqa_result}, \Cref{table:main_qampari_result}, and \Cref{table:main_eli5_result} present the results on ASQA, QAMPARI, and ELI5, respectively.
We have three discovery across the three datasets:
\begin{enumerate}[noitemsep,nolistsep,leftmargin=*]
    \item Snippet+ICLCite performs as the strongest baseline. We hypothesize that for tasks involving grounded generation, preserving original evidence within documents is crucial for citation quality. Abstractive summarization can result in the loss of significant evidence crucial for solving the task. Additionally, accurately extracting consistent answers from multiple samples enhances correctness, yet determining corresponding consistent citations poses a challenge.
    \item Despite Snippet+ICLCite being the strongest baseline, it does not always outperform ICLCite. We observe that with a weaker retriever, Snippet+ICLCite often fails to enhance performance. We conjecture that this is attributed to increased noise in these scenarios. More noisy input lists can lead to a higher likelihood of hallucinations and errors during the preprocessing steps, resulting in degraded performance.
    \item \mymodel{} effectively improves the performance in both answer correctness and citation quality, and the improvement is robust against the choice of retriever. We attribute this robustness to our approach of releasing only high-quality responses in each iteration while continuously exploring new batches of documents.
\end{enumerate}

\subsection{Analysis}
\label{sec:analysis}
We conduct analysis of \mymodel{} on the QAMPARI dataset. All the studies are conducted using GTR as the retriever and text-unicorn as the main LM, except where noted in the table.

\mypar{What if we use larger LM as the verifier LM?}
\Cref{table:small_verifier?} shows how the size of the verifier impacts the task performance. We can see that all verifiers improve the performance, demonstrating the robustness of our framework.
Additionally, the smaller LM variants outperforms the largest LM (tulu-2-70b) in citation quality, and the medium-sized LM achieves the best performance. 
A medium-sized LM combines external knowledge integration and information discernment, resulting in a better verifier performance. 
Comparing with using the large LM itself as verifier, our choice of a smaller LM as the verifier performs better.


\mypar{What if the verifier LM could access more than just the cited documents?}
In our automated verification method, the verifier LM is limited to accessing only the cited documents. Removing this constraint simplifies our verification algorithm to re-sampling with a different LM. The results, shown in \Cref{table:ablating_step_3}, demonstrate a notable decline in performance when such verification design is removed, especially in citation quality. This highlights the essential role and effectiveness of our design of automated verification.

\begin{table}[t!]
\centering
\small
\resizebox{1.0\linewidth}{!}{
\setlength{\tabcolsep}{3.5pt}
\renewcommand{\arraystretch}{1.05}
\begin{tabular}{l|ccccc}
    \toprule
    \multirow{2}{*}{\textbf{Choice of Verifier LM.}}
    & \multicolumn{2}{c}{\textbf{Correctness}} & \multicolumn{2}{c}{\textbf{Citation}}  & \multirow{2}{*}{\textbf{Average}} \\
    \cmidrule{2-5}
    & Prec. & Rec.-5 & Prec. & Rec. & \\
    \midrule
    \multicolumn{6}{c}{\textbf{GPT-3.5-Turbo-1106 as Main LM}} \\
    \midrule
    No verification & 19.23 & 17.32 & 21.75 & 20.77 & 19.77 \\
    \midrule
    Tulu-2-dpo-7b & 25.36 & 17.95 & 24.29 & 23.64 & 22.81 {\scriptsize \color{mygreen}{(+3.04)}}\\
    Tulu-2-dpo-13b (our choice) & \textbf{26.71} & 18.55 & \textbf{25.16} & \textbf{24.39} & \textbf{23.70} {\scriptsize \color{mygreen}{(+3.93)}} \\
    Tulu-2-dpo-70b & 25.61 & \textbf{18.69} & 23.64 & 23.57 & 22.88 {\scriptsize \color{mygreen}{(+3.11)}}\\
    \midrule
    GPT-3.5-Turbo-1106 & 24.39 & 18.36 & 22.94 & 22.30 & 22.00 {\scriptsize \color{mygreen}{(+2.23)}}
    \\
    \midrule
    \multicolumn{6}{c}{\textbf{text-unicorn as Main LM}} \\
    \midrule
    No verification & 26.27 & 21.42 & 25.52 & 25.21 & 24.61 \\
    \midrule
    Tulu-2-dpo-7b & 29.71 & 21.80 & 27.67 & 27.45 & 26.65 {\scriptsize \color{mygreen}{(+2.04)}}\\
    Tulu-2-dpo-13b (our choice) & \textbf{30.28} & 22.79 &\textbf{28.53} & \textbf{28.34} & \textbf{27.49} {\scriptsize \color{mygreen}{(+2.88)}}\\
    Tulu-2-dpo-70b & 28.94 & 22.61 & 27.06 & 26.88 & 26.37 {\scriptsize \color{mygreen}{(+1.76)}}\\
    \midrule
    text-unicorn & 29.73 & \textbf{23.44} & 28.01 & 27.54 & 27.18 {\scriptsize \color{mygreen}{(+2.57)}} \\
    \bottomrule
\end{tabular}
}
\caption{Evaluating the choice of verifier LM. Our investigation focuses on comparing the performance of models of varying sizes and using the main LM itself as the verifier. 
Compared to the largest LM in the same family, a smaller-sized LM yields better performance.
Additionally, our choice of verifier can largely outperform using the main LM itself as verifier.}
\label{table:small_verifier?}
\end{table}

\begin{table}[t!]
\centering
\small
\resizebox{1.0\columnwidth}{!}{
\setlength{\tabcolsep}{3.5pt}
\renewcommand{\arraystretch}{1.03}
\begin{tabular}{l|ccccc}
    \toprule
    \multirow{2}{*}{\textbf{Verifier LM's input document}} & \multicolumn{2}{c}{\textbf{Correctness}} & \multicolumn{2}{c}{\textbf{Citation}}  & \multirow{2}{*}{\textbf{Average}}\\
    & Prec. & Rec.-5 & Prec. & Rec. & \\
    \midrule
    Only cited documents (Ours.) & 30.28 & 22.79 & 28.53 & 28.34 & 27.49 \\
    Same as the main LM & 28.19 & 23.26 & 25.22 & 25.65 & 25.58 \\
    \bottomrule
\end{tabular}
}
\caption{Ablating our design of using only cited documents for verification. We can observe a significant performance drop if we remove the design.}
\vspace{-6pt}
\label{table:ablating_step_3}
\end{table}

\mypar{Further Analysis}
Appendix~\ref{appx:iteration_proceed} details additional studies on model performance and prediction changes across iterations.

%% file: 10-related.tex
\section{Related Work}
\label{sec:related}

\mypar{Evaluation}
Early research focused on evaluating attribution in text generation. \citet{10.1162/coli_a_00486} introduced the ``Attributable to Identified Sources'' (AIS) framework for assessing faithfulness. Subsequent studies developed automatic \cite{honovich-etal-2022-true-evaluating, yue-etal-2023-automatic} and human \cite{bohnet2022attributed} evaluation methods based on AIS. Recent work by \citet{DBLP:conf/emnlp/LiuZL23} evaluated generative search engines that provide citations, and \citet{gao-etal-2023-enabling} proposed ALCE, an automatic benchmark for text generation with citations. In this study, we assess our approach using ALCE.

\mypar{Finetuned LMs} 
Some studies have investigated fine-tuning language models (LMs) for generating cited answers \cite{menick2022teaching, nakano2021webgpt}. Similarly, \citet{ye2023effective} employed adaptation approach for fine-tuning. These methods required training and could be be susceptible to generalization issues.

\mypar{Retrieval-based Methods} 
\citet{he2022rethinking, gao-etal-2023-rarr} used post-editing to ensure text consistency by retrieving relevant documents. \citet{gao-etal-2023-enabling} explored methods like document summarization and LLM-enabled searches for citation improvement, yet lacked verification for answer validation. \citet{LLatrieval2023Li} utilized an LLM as a verifier for document relevance but didn't employ the answer to verify the correct grounding.

\mypar{Self-reflection} 
Prompting LLMs to self-reflect on their answers has been shown to improve factuality \cite{ji-etal-2023-towards}. \citet{asai2023self} employed this concept, enhancing LM quality and factuality via retrieval and self-reflection by training special tokens. \mymodel{} outperforms this method without the need of training.

%% file: 06-conclusion.tex
\section{Conclusion}
\label{sec:conclusion}
In this paper, we introduce \mymodel{}, a novel verification approach for grounded generation. We observe that while larger LMs excel at identifying relevant materials, they tend to rely excessively on internal parametric memory. Conversely, smaller LMs are adept at processing focused information.  \mymodel{} leverages these complementary strengths to offer a fresh perpective on robust and scalable solution for verification in grounded generation.


\section*{Limitation}
\label{sec:limit}
We acknowledge the limitations of \mymodel{} from the following aspects to inspire future research opportunities in the field of grounded generation.

Firstly, as a postprocessing technique, our method introduces additional latency in generating the final answer. As a remedy, we can set the maximum iteration $T$ smaller. From \Cref{fig:iter_performance} in the appendix, we have shown that only even a single iteration of our correction process significantly enhances performance, yet latency remains an unavoidable factor.

Moreover, unlike preprocessing approaches depicted in \Cref{fig:teaser}(b), which can reduce input token consumption for the final LLM, our method necessitates that the LLM initially processes all documents, leading to a higher cost for token usage.

Lastly, despite the considerable advancements \mymodel{} has made across datasets, the instances that pass our verification process are still not flawless. Given that both the answer candidate and the verifier output are outcomes from LMs, there is an inevitable risk of both models producing hallucinations simultaneously. 

We hope future works can leverage the idea and insights from \mymodel{} to advance the development of more robust grounded generation with low latency and reduced token costs.

\section*{Broader Considerations}
As a method that directly apply LLMs, \mymodel{} inherits all potential risks associated with LLMs, including but not limited to unethical outputs, toxicity, and biases~\cite{DBLP:conf/fat/BenderGMS21, DBLP:journals/corr/abs-2401-10019, DBLP:journals/corr/abs-2309-00770}. 
Our qualitative assessment of \mymodel{}, conducted across several samples from three datasets, indicates that LLMs generally adhere to instructions and generate responses relevant to the content of provided documents. However, we strongly advise conducting a comprehensive evaluation of these potential issues before deploying \mymodel{} in practical settings.

%% file: 99-appendix.tex
\clearpage
\appendix

\section{Further Analysis}
\label{appx:iteration_proceed}
In this section, we focus on analysis when iteration of \mymodel{} proceed.

\begin{figure*}[h]
\centering 
\includegraphics[width=0.85\textwidth]{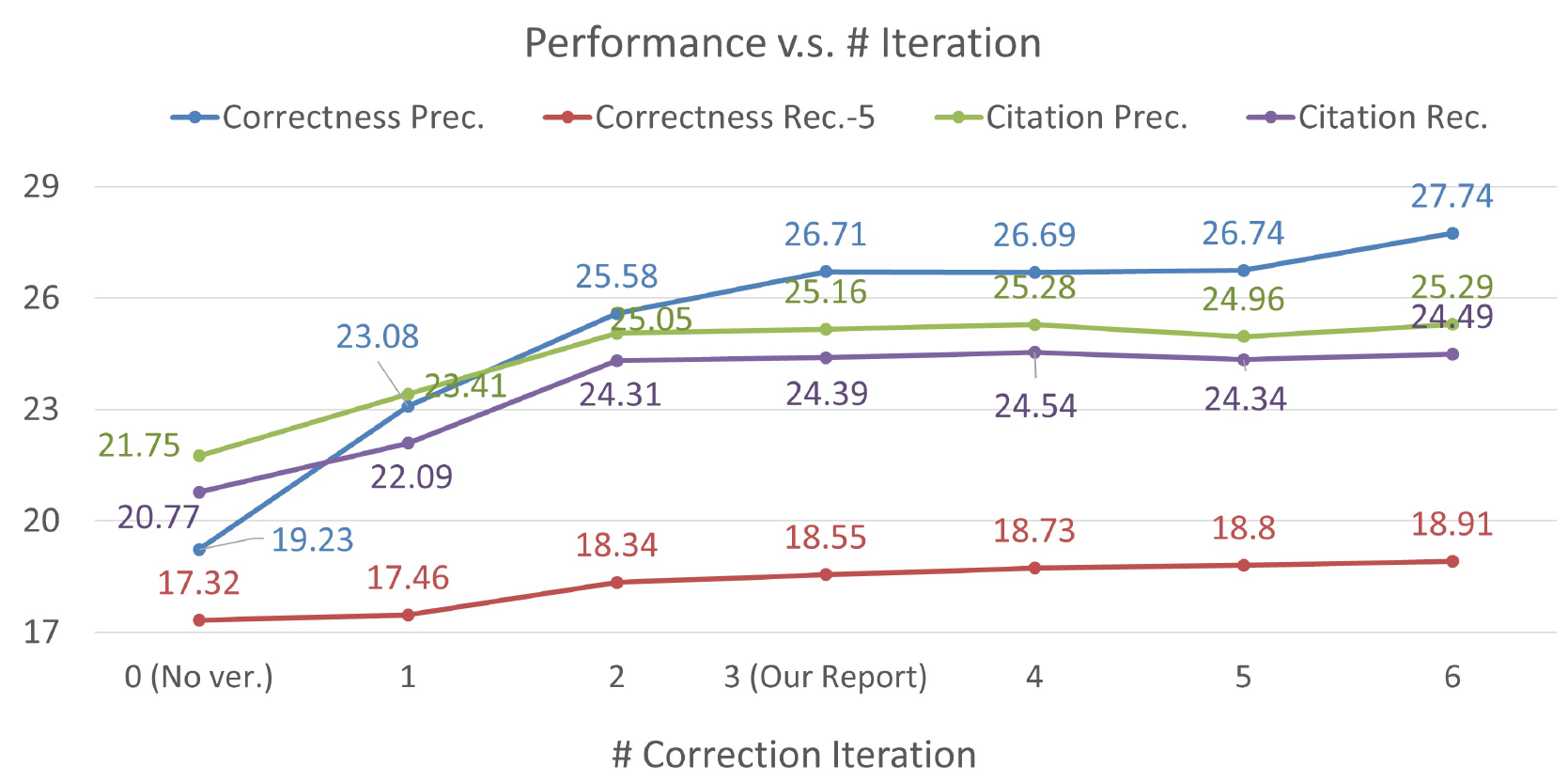}
\caption{The study examines the iterative performance improvements on the QAMPARI dataset. We use GPT-3.5-Turbo as the main LM for this running study.} 
\label{fig:iter_performance} 
\end{figure*}

\mypar{How does the model's performance improve as iterations proceed?} 
\Cref{fig:iter_performance} illustrates the outcomes of terminating the process from iteration $0$ through $6$.
As demonstrated in the figure, the precision for correctness consistently improve with each iteration of CaLM as our framework only allows high quality final output.
On the other hand, our design of updating the input document list contributes to a consistent rise in correctness recall.

However, we observe that performance tends to plateau after the third iteration, with subsequent iterations yielding diminishing returns. 
Extending the maximum iterations to six produced only a marginal average improvement of 0.4 compared to the third iteration, while significantly increasing the computational cost in terms of API calls.

We believe \mymodel{}’s iterative nature is a key strength, allowing for continuous improvement. However, our findings suggest that two to three iterations offer substantial gains with minimal computational overhead. This demonstrates the framework’s efficiency and practicality for real-world applications.

\mypar{Case study on \mymodel{}'s correction.}
We present two case studies to demonstrate that by looping through our verification design, more accurate evidence can be found and more accurate responses can be generated. \Cref{fig:case_asqa} is based on the ASQA dataset, and \Cref{fig:case_qampari} utilizes the QAMPARI dataset. 

From the example of \Cref{fig:case_asqa}, we can see that the small LM, serving as a verifier, when given access to cleaner input document sets, is capable of identifying overlooked information by the main LM. This detection initiates iterative correction processes in subsequent rounds.

The example in \Cref{fig:case_qampari} demonstrates that (1) \mymodel{} finds more convincing evidence documents in later rounds, and (2) \mymodel{} catches citation errors through verification.

\begin{figure*}[h!]
\begin{lstlisting}
> Round 1:

>> Main LM:
>>> Access to:
Document [1](Title: Don't Tell Me What to Do): "Don't Tell Me What to Do" is a song written by Harlan Howard and Max D. Barnes, and recorded by American country music artist Pam Tillis. It was released in December 1990 as the first single from the album "Put Yourself in My Place". The song reached number 5 on the "Billboard" Hot Country Singles & Tracks chart. Marty Stuart recorded this song under the title "I'll Love You Forever (If I Want To)" in 1988

Document [2](Title: Don't (Ed Sheeran song)): (Omitted)

Document [3](Title: Don't Speak): (Omitted)

Document [4](Title: Don't Tell Me What to Do): for his "Let There Be Country" album, although the album was not released until 1992. Don't Tell Me What to Do "for Baby Animals' song of the same name, see "Don't Tell Me What to Do (Baby Animals song)". "Don't Tell Me What to Do" is a song written by Harlan Howard and Max D. Barnes, and recorded by American country music artist Pam Tillis. It was released in December 1990 as the first single from the album ...(Omitted)

Document [5](Title: Pam Tillis): Pam Tillis Pamela Yvonne Tillis (born July 24, 1957) is an American country music singer-songwriter and actress. She is the daughter of the late country music singer Mel Tillis and Doris Tillis. ...(Omitted)... By 1991, she had signed to Arista Nashville, entering Top 40 on Hot Country Songs for the first time with "Don't Tell Me What to Do", the first of five

>>> Response: 
The song "Don't Tell Me What to Do" was written by Harlan Howard and Max D. Barnes and recorded by American country music artist Pam Tillis, reaching number 5 on the "Billboard" Hot Country Singles & Tracks chart [1][4][5].

>> Verifier LM:
>>> Access to: Document 1,4,5
>>> Response: 
Pam Tillis recorded the song "Don't Tell Me What to Do" [5], which reached the Top 40 of Hot Country Songs in 1991 [5]. This song was also written by Harlan Howard and Max D. Barnes, and Marty Stuart recorded a version of this song [1][4].

> Round 2:

>> Main LM:
>>> Access to:
Document [1](Title: Don't Tell Me What to Do): "Don't Tell Me What to Do" is a song written by Harlan Howard and Max D. Barnes, and recorded by American country music artist Pam Tillis. It was released in December 1990 as the first single from the album "Put Yourself in My Place". The song reached number 5 on the "Billboard" Hot Country Singles & Tracks chart. Marty Stuart recorded this song under the title "I'll Love You Forever (If I Want To)" in 1988

Document [4](Title: Don't Tell Me What to Do): for his "Let There Be Country" album, although the album was not released until 1992. Don't Tell Me What to Do "for Baby Animals' song of the same name, see "Don't Tell Me What to Do (Baby Animals song)". "Don't Tell Me What to Do" is a song written by Harlan Howard and Max D. Barnes, and recorded by American country music artist Pam Tillis. It was released in December 1990 as the first single from the album ...(Omitted)

Document [5](Title: Pam Tillis): Pam Tillis Pamela Yvonne Tillis (born July 24, 1957) is an American country music singer-songwriter and actress. She is the daughter of the late country music singer Mel Tillis and Doris Tillis. ...(Omitted)... By 1991, she had signed to Arista Nashville, entering Top 40 on Hot Country Songs for the first time with "Don't Tell Me What to Do", the first of five

Document [6](Title: Don't Speak): (Omitted)

Document [7](Title: Don't (Elvis Presley song)): (Omitted)

>>> Response: 
The song "Don't Tell Me What to Do" was written by Harlan Howard and Max D. Barnes and recorded by American country music artist Pam Tillis, reaching number 5 on the "Billboard" Hot Country Singles & Tracks chart [1][4][5]. Additionally, Marty Stuart also recorded this song under the title "I'll Love You Forever (If I Want To)" in 1988 [1].

(Omit the rest of the steps for clarity.)

\end{lstlisting}
\caption{A case study of \mymodel{} on ASQA dataset. The question is \textit{``Who sings don't tell me what to do?''} and all reference short answers are \textit{``Pam Tillis''}, \textit{``Marty Stuart''}, and \textit{``Baby Animals''}.}
\label{fig:case_asqa}
\end{figure*}

\begin{figure*}[h!]
\begin{lstlisting}
> Round 1:

>> Main LM:
>>> Access to:
Document [1](Title: James Newton Howard):  (Omitted)

Document [2](Title: John Carpenter): in October 2018. The film acts as a direct sequel to Carpenter's original film, ignoring the continuity of all other previous films. It is his first direct involvement with the franchise since 1982's ". Carpenter's films are characterized by minimalist lighting and photography, static cameras, use of steadicam, and distinctive synthesized scores (usually self-composed). With the exception of "Someone's Watching Me!", "Elvis", "The Thing", "Starman", "Memoirs of an Invisible Man", and "The Ward", he has scored all of his films (though some are collaborations), most famously the themes from "Halloween" and "Assault on Precinct 13". His music is generally synthesized

Document [3](Title: Thomas Newman):  (Omitted)

Document [4](Title: John Carpenter): a score by Ennio Morricone and a cast including young actor Kurt Russell and respected character actors such as Wilford Brimley, Richard Dysart, Charles Hallahan, Keith David, and Richard Masur. "The Thing" was distributed by Universal Pictures. Although Carpenter's film used the same source material as the 1951 Howard Hawks film, "The Thing from Another World", it is more faithful to the John W. Campbell, Jr. novella, "Who Goes There?", upon which both films were based. Moreover, unlike the Hawks film, "The Thing" was part of what Carpenter later called his "Apocalypse Trilogy," a trio of films ("The Thing", "Prince

Document [5](Title: Alan Howarth (composer)): (Omitted)

>>> Response: 
Assault on Precinct 13 [2], Halloween [2], The Thing [4].

>> Verifier LM:
>>> Access to: Document 2,4
>>> Response: 
Assault on Precinct 13 [2].

> Round 2:

>> Main LM:
>>> Access to:
Document [2](Title: John Carpenter): in October 2018. The film acts as a direct sequel to Carpenter's original film, ignoring the continuity of all other previous films. It is his first direct involvement with the franchise since 1982's ". Carpenter's films are characterized by minimalist lighting and photography, static cameras, use of steadicam, and distinctive synthesized scores (usually self-composed). With the exception of "Someone's Watching Me!", "Elvis", "The Thing", "Starman", "Memoirs of an Invisible Man", and "The Ward", he has scored all of his films (though some are collaborations), most famously the themes from "Halloween" and "Assault on Precinct 13". His music is generally synthesized

Document [4](Title: John Carpenter): a score by Ennio Morricone and a cast including young actor Kurt Russell and respected character actors such as Wilford Brimley, Richard Dysart, Charles Hallahan, Keith David, and Richard Masur. "The Thing" was distributed by Universal Pictures. Although Carpenter's film used the same source material as the 1951 Howard Hawks film, "The Thing from Another World", it is more faithful to the John W. Campbell, Jr. novella, "Who Goes There?", upon which both films were based. Moreover, unlike the Hawks film, "The Thing" was part of what Carpenter later called his "Apocalypse Trilogy," a trio of films ("The Thing", "Prince

Document [6](Title: John Carpenter): classics that Carpenter has directed include "Dark Star" (1974), "Assault on Precinct 13" (1976), "The Thing" (1982), "Christine" (1983), "Big Trouble in Little China" (1986), "Prince of Darkness" (1987), "They Live" (1988), and "In the Mouth of Madness" (1995). He returned to the "Halloween" franchise as both composer and executive producer for the horror sequel "Halloween" (2018). Carpenter is also notable for having composed or co-composed most of his films' music; some of them are now well-known, with the main theme of "Halloween" being considered a part of popular culture. He won a Saturn Award for Best Music for the

Document [7](Title: Donald Rubinstein): (Omitted)

Document [8](Title: Nellee Hooper): (Omitted)

>>> Response: 
Halloween [6], Dark Star [6], Assault on Precinct 13 [6], The Thing [6], Christine [6], Big Trouble in Little China [6], Prince of Darkness [6], They Live [6], In the Mouth of Madness [6]

>> Verifier LM:
>>> Access to: Document 6
>>> Response: 
Halloween [6], Dark Star [6], The Thing [6], Christine [6], Big Trouble in Little China [6], Prince of Darkness [6], They Live [6], In the Mouth of Madness [6]

(Omit the rest of the steps for clarity.)

\end{lstlisting}
\caption{A case study of \mymodel{} on QAMPARI dataset. The question is \textit{``Which movie did John Carpenter direct for which he also composed the music?''} and all reference answers are \textit{``Vampires'', ``In the Mouth of Madness'', ``Assault on Precinct 13'', ``Dark Star'', ``Big Trouble in Little China'', ``They Live'', ``Halloween'', ``Escape from New York'', ``Prince of Darkness'', ``Ghosts of Mars'', ``The Fog'', ``Chevil'', ``Village of the Damned''}.}
\label{fig:case_qampari}
\end{figure*}

\section{Used prompts}
\label{appx:prompt}
In this section, we list the prompts we use for our experiment.
\subsection{Prompts for ASQA}
Two different prompt sets are used for the ASQA dataset. \Cref{fig:prompt-asqa-main} shows the prompts we used for the LM to conduct grounded generation, which mainly follow the prompt used in \cite{gao-etal-2023-enabling} with two shot examples. We design our own prompt for the main model to perform correction. The prompt is detailed in \Cref{fig:prompt-asqa-correct}, which use 1-shot example.

\begin{figure*}
\begin{lstlisting}
Instruction: Write an accurate, engaging, and concise answer for the given question using only the provided search results (some of which might be irrelevant) and cite them properly. Use an unbiased and journalistic tone. Always cite for factual claims. When citing several search results, use [1][2][3]. Cite at least one and no more than three documents per sentence. If multiple documents support the sentence, only cite a minimum sufficient subset.

Question: {EXAMPLE 1}

Document [1](Title: {TITLE_FOR_DOC_1}): {DOC_1}
Document [2](Title: {TITLE_FOR_DOC_2}): {DOC_2}
Document [3](Title: {TITLE_FOR_DOC_3}): {DOC_3}
Document [4](Title: {TITLE_FOR_DOC_4}): {DOC_4}
Document [5](Title: {TITLE_FOR_DOC_5}): {DOC_5}

Answer:  {ANSWER_FOR_EXAMPLE 1}

Instruction: Write an accurate, engaging, and concise answer for the given question using only the provided search results (some of which might be irrelevant) and cite them properly. Use an unbiased and journalistic tone. Always cite for factual claims. When citing several search results, use [1][2][3]. Cite at least one and no more than three documents per sentence. If multiple documents support the sentence, only cite a minimum sufficient subset.

Question: {EXAMPLE 2}

Document [1](Title: {TITLE_FOR_DOC_1}): {DOC_1}
Document [2](Title: {TITLE_FOR_DOC_2}): {DOC_2}
Document [3](Title: {TITLE_FOR_DOC_3}): {DOC_3}
Document [4](Title: {TITLE_FOR_DOC_4}): {DOC_4}
Document [5](Title: {TITLE_FOR_DOC_5}): {DOC_5}

Answer:  {ANSWER_FOR_EXAMPLE 2}

Instruction: Write an accurate, engaging, and concise answer for the given question using only the provided search results (some of which might be irrelevant) and cite them properly. Use an unbiased and journalistic tone. Always cite for factual claims. When citing several search results, use [1][2][3]. Cite at least one and no more than three documents per sentence. If multiple documents support the sentence, only cite a minimum sufficient subset.

Question: {REAL QUERY}

Document [1](Title: {TITLE_FOR_DOC_1}): {DOC_1}
Document [2](Title: {TITLE_FOR_DOC_2}): {DOC_2}
Document [3](Title: {TITLE_FOR_DOC_3}): {DOC_3}
Document [4](Title: {TITLE_FOR_DOC_4}): {DOC_4}
Document [5](Title: {TITLE_FOR_DOC_5}): {DOC_5}

Answer: 
\end{lstlisting}
\caption{Prompt for the LM to conduct the grounded generation on ASQA dataset.}
\label{fig:prompt-asqa-main}
\end{figure*}

\begin{figure*}
\begin{lstlisting}
Instruction: Provide a concise response to the question by analyzing relevant search results (some of which might be irrelevant), and cite useful resources using [1][2][3] format. Use an unbiased and journalistic tone, ensuring facts are presented clearly based on the search documents. Cite at least one and no more than three documents per sentence. Additionally, you will be provided with an incomplete draft solution, which is based on the first {SIZE_OF_VERIFIED_DOCS} search results and might contain citation inaccuracies. Therefore, it might not capture all conceivable responses to the question. Your role is to assess the draft's comprehensiveness as well as its correctness, and then update the solution to encapsulate all possible answers according to the search documents. Provide your answer after "Corrected Answer:", and ensure each sentence is supported by citations from one to three sources.

Question: {EXAMPLE 1}

Document [1](Title: {TITLE_FOR_DOC_1}): {DOC_1}
Document [2](Title: {TITLE_FOR_DOC_2}): {DOC_2}
Document [3](Title: {TITLE_FOR_DOC_3}): {DOC_3}
Document [4](Title: {TITLE_FOR_DOC_4}): {DOC_4}
Document [5](Title: {TITLE_FOR_DOC_5}): {DOC_5}

Drafted Solution: {DRAFT SOLUTION FOR EXAMPLE 1}

Corrected Answer:  {ANSWER_FOR_EXAMPLE 1}

Instruction: Provide a concise response to the question by analyzing relevant search results (some of which might be irrelevant), and cite useful resources using [1][2][3] format. Use an unbiased and journalistic tone, ensuring facts are presented clearly based on the search documents. Cite at least one and no more than three documents per sentence. Additionally, you will be provided with an incomplete draft solution, which is based on the first {SIZE_OF_VERIFIED_DOCS} search results and might contain citation inaccuracies. Therefore, it might not capture all conceivable responses to the question. Your role is to assess the draft's comprehensiveness as well as its correctness, and then update the solution to encapsulate all possible answers according to the search documents. Provide your answer after "Corrected Answer:", and ensure each sentence is supported by citations from one to three sources.

Question: {QUERY}

Document [1](Title: {TITLE_FOR_DOC_1}): {DOC_1}
Document [2](Title: {TITLE_FOR_DOC_2}): {DOC_2}
Document [3](Title: {TITLE_FOR_DOC_3}): {DOC_3}
Document [4](Title: {TITLE_FOR_DOC_4}): {DOC_4}
Document [5](Title: {TITLE_FOR_DOC_5}): {DOC_5}

Drafted Solution: {DRAFT SOLUTION FOR QUERY}

Corrected Answer:
\end{lstlisting}
\caption{Prompt for the main LM to conduct correction on ASQA dataset.}
\label{fig:prompt-asqa-correct}
\end{figure*}

\subsection{Prompts for QAMPARI}
Two different prompt sets are used for the QAMPRI dataset. \Cref{fig:prompt-qampari-main} shows the prompts we used for the LM to conduct grounded generation, which mainly follow the prompt used in \cite{gao-etal-2023-enabling} with two shot examples. We design our own prompt for the main model to perform correction. The prompt is detailed in \Cref{fig:prompt-qampari-correct}, which use 1-shot example.

\begin{figure*}
\begin{lstlisting}
Instruction: Provide a list of accurate answers for the given question using only the provided search results (some of which might be irrelevant) and cite them properly using [1][2][3]. Always cite one and only one document for each answer. Separate answers by commas. For questions that have more than 5 answers, write at least 5 answers.

Question: {EXAMPLE 1}

Document [1](Title: {TITLE_FOR_DOC_1}): {DOC_1}
Document [2](Title: {TITLE_FOR_DOC_2}): {DOC_2}
Document [3](Title: {TITLE_FOR_DOC_3}): {DOC_3}
Document [4](Title: {TITLE_FOR_DOC_4}): {DOC_4}
Document [5](Title: {TITLE_FOR_DOC_5}): {DOC_5}

Answer:  {ANSWER_FOR_EXAMPLE 1}

Instruction: Provide a list of accurate answers for the given question using only the provided search results (some of which might be irrelevant) and cite them properly using [1][2][3]. Always cite one and only one document for each answer. Separate answers by commas. For questions that have more than 5 answers, write at least 5 answers.

Question: {EXAMPLE 2}

Document [1](Title: {TITLE_FOR_DOC_1}): {DOC_1}
Document [2](Title: {TITLE_FOR_DOC_2}): {DOC_2}
Document [3](Title: {TITLE_FOR_DOC_3}): {DOC_3}
Document [4](Title: {TITLE_FOR_DOC_4}): {DOC_4}
Document [5](Title: {TITLE_FOR_DOC_5}): {DOC_5}

Answer:  {ANSWER_FOR_EXAMPLE 2}

Instruction: Provide a list of accurate answers for the given question using only the provided search results (some of which might be irrelevant) and cite them properly using [1][2][3]. Always cite one and only one document for each answer. Separate answers by commas. For questions that have more than 5 answers, write at least 5 answers.

Question: {REAL QUERY}

Document [1](Title: {TITLE_FOR_DOC_1}): {DOC_1}
Document [2](Title: {TITLE_FOR_DOC_2}): {DOC_2}
Document [3](Title: {TITLE_FOR_DOC_3}): {DOC_3}
Document [4](Title: {TITLE_FOR_DOC_4}): {DOC_4}
Document [5](Title: {TITLE_FOR_DOC_5}): {DOC_5}

Answer: 
\end{lstlisting}
\caption{Prompt for the LM to conduct the grounded generation on QAMPARI dataset.}
\label{fig:prompt-qampari-main}
\end{figure*}

\begin{figure*}
\begin{lstlisting}
Instruction: Provide a list of accurate answers for the given question using only the provided search results (some of which might be irrelevant) and cite them properly using [1][2][3]. Always cite one and only one document for each answer. Separate answers by commas. For questions that have more than 5 answers, write at least 5 answers. Additionally, you will receive an incorrect draft or incomplete draft solution, which is based on the first {SIZE_OF_VERIFIED_DOCS} search results. Hence, it may not encompass all potential answers. Your task is to check this solution and modify it to comprehensively answer the question, using information from the additional provided search results. If the draft is missing, start from scratch, basing your answer on the search outcomes. Please begin by concisely explain your evaluation of the draft solution within 150 words. Then, present your conclusive response under the heading "Final Answer:"

Question: {EXAMPLE 1}

Document [1](Title: {TITLE_FOR_DOC_1}): {DOC_1}
Document [2](Title: {TITLE_FOR_DOC_2}): {DOC_2}
Document [3](Title: {TITLE_FOR_DOC_3}): {DOC_3}
Document [4](Title: {TITLE_FOR_DOC_4}): {DOC_4}
Document [5](Title: {TITLE_FOR_DOC_5}): {DOC_5}

Drafted Solution: {DRAFT SOLUTION FOR EXAMPLE 1}

Corrected Answer:  {ANSWER_FOR_EXAMPLE 1}

Instruction: Provide a list of accurate answers for the given question using only the provided search results (some of which might be irrelevant) and cite them properly using [1][2][3]. Always cite one and only one document for each answer. Separate answers by commas. For questions that have more than 5 answers, write at least 5 answers. Additionally, you will receive an incorrect draft or incomplete draft solution, which is based on the first {SIZE_OF_VERIFIED_DOCS} search results. Hence, it may not encompass all potential answers. Your task is to check this solution and modify it to comprehensively answer the question, using information from the additional provided search results. If the draft is missing, start from scratch, basing your answer on the search outcomes. Please begin by concisely explain your evaluation of the draft solution within 150 words. Then, present your conclusive response under the heading "Final Answer:"

Question: {QUERY}

Document [1](Title: {TITLE_FOR_DOC_1}): {DOC_1}
Document [2](Title: {TITLE_FOR_DOC_2}): {DOC_2}
Document [3](Title: {TITLE_FOR_DOC_3}): {DOC_3}
Document [4](Title: {TITLE_FOR_DOC_4}): {DOC_4}
Document [5](Title: {TITLE_FOR_DOC_5}): {DOC_5}

Drafted Solution: {DRAFT SOLUTION FOR QUERY}

Corrected Answer:
\end{lstlisting}
\caption{Prompt for the main LM to conduct correction on QAMPARI dataset.}
\label{fig:prompt-qampari-correct}
\end{figure*}

\subsection{Prompts for ELI5}
Two different prompt sets are used for the ELI5 dataset. \Cref{fig:prompt-eli5-main} shows the prompts we used for the LM to conduct grounded generation, which mainly follow the prompt used in \cite{gao-etal-2023-enabling} with two shot examples. We design our own prompt for the main model to perform correction. The prompt is detailed in \Cref{fig:prompt-eli5-correct}, which use 1-shot example.

\begin{figure*}
\begin{lstlisting}
Instruction: Write an accurate, engaging, and concise answer for the given question using only the provided search results (some of which might be irrelevant) and cite them properly. Use an unbiased and journalistic tone. Always cite for any factual claim. When citing several search results, use [1][2][3]. Cite at least one document and at most three documents in each sentence. If multiple documents support the sentence, only cite a minimum sufficient subset of the documents.

Question: {EXAMPLE 1}

Document [1](Title: {TITLE_FOR_DOC_1}): {DOC_1}
Document [2](Title: {TITLE_FOR_DOC_2}): {DOC_2}
Document [3](Title: {TITLE_FOR_DOC_3}): {DOC_3}
Document [4](Title: {TITLE_FOR_DOC_4}): {DOC_4}
Document [5](Title: {TITLE_FOR_DOC_5}): {DOC_5}

Answer:  {ANSWER_FOR_EXAMPLE 1}

Instruction: Write an accurate, engaging, and concise answer for the given question using only the provided search results (some of which might be irrelevant) and cite them properly. Use an unbiased and journalistic tone. Always cite for any factual claim. When citing several search results, use [1][2][3]. Cite at least one document and at most three documents in each sentence. If multiple documents support the sentence, only cite a minimum sufficient subset of the documents.

Question: {EXAMPLE 2}

Document [1](Title: {TITLE_FOR_DOC_1}): {DOC_1}
Document [2](Title: {TITLE_FOR_DOC_2}): {DOC_2}
Document [3](Title: {TITLE_FOR_DOC_3}): {DOC_3}
Document [4](Title: {TITLE_FOR_DOC_4}): {DOC_4}
Document [5](Title: {TITLE_FOR_DOC_5}): {DOC_5}

Answer:  {ANSWER_FOR_EXAMPLE 2}

Instruction: Write an accurate, engaging, and concise answer for the given question using only the provided search results (some of which might be irrelevant) and cite them properly. Use an unbiased and journalistic tone. Always cite for any factual claim. When citing several search results, use [1][2][3]. Cite at least one document and at most three documents in each sentence. If multiple documents support the sentence, only cite a minimum sufficient subset of the documents.

Question: {REAL QUERY}

Document [1](Title: {TITLE_FOR_DOC_1}): {DOC_1}
Document [2](Title: {TITLE_FOR_DOC_2}): {DOC_2}
Document [3](Title: {TITLE_FOR_DOC_3}): {DOC_3}
Document [4](Title: {TITLE_FOR_DOC_4}): {DOC_4}
Document [5](Title: {TITLE_FOR_DOC_5}): {DOC_5}

Answer: 
\end{lstlisting}
\caption{Prompt for the LM to conduct the grounded generation on ELI5 dataset.}
\label{fig:prompt-eli5-main}
\end{figure*}

\begin{figure*}
\begin{lstlisting}
Instruction: Provide a succint response to the question by analyzing relevant search results (some of which might be irrelevant), and cite these using [1][2][3] format. Limit citations to one to three per sentence, and only cite a necessary subset if multiple sources corroborate a point. Additionally, you will be provided with an incomplete draft solution, which is based on the first {SIZE_OF_VERIFIED_DOCS} search results and might contain citation inaccuracies. Therefore, it might not capture all conceivable responses to the question. Your role is to assess the draft's comprehensiveness as well as its correctness of citations, and then update the solution to encapsulate all possible answers according to the search documents. Provide your answer after "Corrected Answer:", and ensure each sentence is supported by citations from one to three sources.

Question: {EXAMPLE 1}

Document [1](Title: {TITLE_FOR_DOC_1}): {DOC_1}
Document [2](Title: {TITLE_FOR_DOC_2}): {DOC_2}
Document [3](Title: {TITLE_FOR_DOC_3}): {DOC_3}
Document [4](Title: {TITLE_FOR_DOC_4}): {DOC_4}
Document [5](Title: {TITLE_FOR_DOC_5}): {DOC_5}

Drafted Solution: {DRAFT SOLUTION FOR EXAMPLE 1}

Corrected Answer:  {ANSWER_FOR_EXAMPLE 1}

Instruction: Provide a succint response to the question by analyzing relevant search results (some of which might be irrelevant), and cite these using [1][2][3] format. Limit citations to one to three per sentence, and only cite a necessary subset if multiple sources corroborate a point. Additionally, you will be provided with an incomplete draft solution, which is based on the first {SIZE_OF_VERIFIED_DOCS} search results and might contain citation inaccuracies. Therefore, it might not capture all conceivable responses to the question. Your role is to assess the draft's comprehensiveness as well as its correctness of citations, and then update the solution to encapsulate all possible answers according to the search documents. Provide your answer after "Corrected Answer:", and ensure each sentence is supported by citations from one to three sources.

Question: {QUERY}

Document [1](Title: {TITLE_FOR_DOC_1}): {DOC_1}
Document [2](Title: {TITLE_FOR_DOC_2}): {DOC_2}
Document [3](Title: {TITLE_FOR_DOC_3}): {DOC_3}
Document [4](Title: {TITLE_FOR_DOC_4}): {DOC_4}
Document [5](Title: {TITLE_FOR_DOC_5}): {DOC_5}

Drafted Solution: {DRAFT SOLUTION FOR QUERY}

Corrected Answer:
\end{lstlisting}
\caption{Prompt for the main LM to conduct correction on ELI5 dataset.}
\label{fig:prompt-eli5-correct}
\end{figure*}

\section{Implementation Details}
\label{appx:implement}
For all experiments with public available models, we use vLLM framework for inference~\cite{kwon2023efficient}. 
We operate vLLM on our machine with 16 NVIDIA-A100-40GB GPU.
For experiments with GPT-3.5-Turbo-1106, we use the official API~\footnote{https://platform.openai.com/}. For experiment with text-unicorn, we use Google-Cloud vertex API~\footnote{https://cloud.google.com/vertex-ai/docs/reference/rest}.

\section{Dataset and Evaluation Tool}
We use the artifacts provided by \citet{gao-etal-2023-enabling}. The dataset and corresponding evaluation code is under MIT licence~\footnote{https://github.com/princeton-nlp/ALCE/tree/main}. We do not change any of the provided data and maintain consistent with their intended use.
